\documentclass[runningheads,orivec]{llncs}
\usepackage[T1]{fontenc}
\usepackage{graphicx}

\usepackage{amsfonts}
\usepackage{amssymb}
\usepackage{amsmath}
\usepackage{mathtools}
\usepackage{booktabs}
\usepackage{multirow}
\usepackage{algorithm2e}
\usepackage{wrapfig}
\usepackage{subcaption}
\usepackage{enumitem}%
\SetKwComment{Comment}{// }{}

\usepackage{xcolor}

\newcommand{\para}[1]{\phantom{}\newline\noindent\textbf{#1.}}

\begin{document}
\title{Modeling Behavioral Patterns in News Recommendations Using Fuzzy Neural Networks}
\titlerunning{News Recommendations Using Fuzzy Neural Networks}

\author{Kevin Innerebner\orcidID{0009-0001-8556-5371} \and
Stephan Bartl\orcidID{0009-0002-6645-5617} \and
Markus Reiter-Haas\orcidID{0000-0001-9852-8206} \and
Elisabeth Lex\orcidID{0000-0001-5293-2967}
}

\authorrunning{Innerebner et al.}
\institute{Institute of Human-Centred Computing, Graz University of Technology, Austria}
\maketitle              %
\begin{abstract}

News recommender systems are increasingly driven by black-box models, offering little transparency for editorial decision-making. 
In this work, we introduce a transparent recommender system that uses fuzzy neural networks to learn human-readable rules from behavioral data for predicting article clicks. 
By extracting the rules at configurable thresholds, 
we can control rule complexity and thus, the level of interpretability. 
We evaluate our approach on two publicly available news datasets (i.e., MIND and EB-NeRD) and show that we can accurately predict click behavior compared to several established baselines, while learning human-readable rules. 
Furthermore, we show that the learned rules reveal news consumption patterns, enabling editors to align content curation goals with target audience behavior. 

\end{abstract}

\begin{figure}
    \centering
    \includegraphics[width=1\linewidth]{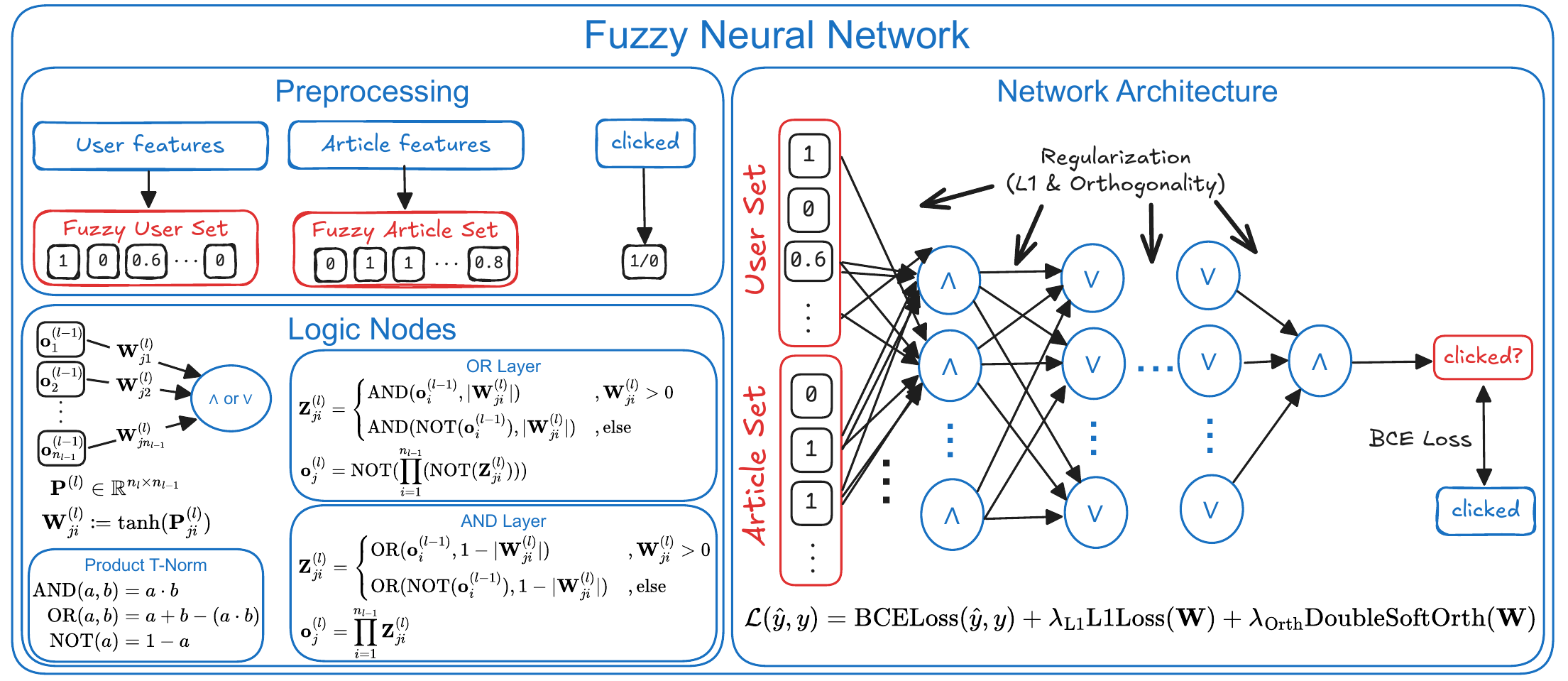}
    \caption{Architecture of our model design. User and article features are converted to fuzzy sets, serving as the input. Our model features multiple layers of logic nodes that perform product t-norm operations ($\land$ or $\lor$) on their inputs. Each input to the logic node is weighted so that high weights result in a strong influence of the input, while low weights mean the input has little influence on the computation. Inputs are negated ($\lnot$) when the corresponding weight is negative.}
    \label{fig:architecture}
\end{figure}

\section{Introduction}

News organizations increasingly rely on recommender systems to recommend articles to readers. While these systems can improve engagement, they are typically built on deep learning models that provide limited transparency or control for editors. This is particularly problematic in domains where editorial gatekeeping is central to combating disinformation and supporting an informed public~\cite{olsen2022gatekeepers}. Editors increasingly face a dual challenge, i.e., fulfilling their traditional editorial responsibilities while auditing and steering algorithmic news curation, often with insufficient insight into why certain articles are recommended~\cite{moller2022recommended,peukert2024editor}.

\noindent Existing news recommender systems largely prioritize predictive performance~\cite{raza2022news}. However, the news domain has unique characteristics, such as fast-changing temporal dynamics in news consumption and the risk of polarization~\cite{bentley2019understanding} that require not only accurate predictions but interpretable insight into reader behavior~\cite{bentley2019understanding}. Hybrid approaches that combine algorithms with editorially curated picks offer a partial remedy~\cite{klimashevskaia2024empowering}, but they do not explain the behavioral patterns underlying model decisions. We argue what editors need are news recommender techniques that are transparent by design, transforming behavioral patterns into human-readable rules that can be analyzed, communicated, and, where needed, adjusted. While recommender systems can be built purely with fuzzy logic, for instance, fuzzy inference to predict article similarity~\cite{adnan2014content} or category interest~\cite{manoharan2020intelligent}, the rule bases are typically hand-crafted from domain knowledge~\cite{adnan2014content} or mined with heuristics. A common neuro-symbolic alternative combines fuzzy logic with neural networks. In the Wang-Mendel pipeline~\cite{WangMendelFuzzy1992}, inputs are first fuzzified and IF-THEN rules are mined from input-output pairs; these fuzzy sets/rules can then be coupled to neural architectures in several ways: (i) encode user preferences as fuzzy sets and feed them into a network~\cite{rutkowskiContentBasedRecommendationSystem2018}, optionally tuning memberships via backpropagation~\cite{dengNovelFuzzyNeural2024}; (ii) map inputs to fuzzy atoms with an MLP and apply a fixed logic circuit on top~\cite{ochoaMedicalRecommenderSystems2021}; (iii) use fuzzy rules to mitigate sparsity by predicting missing (multi-criteria) ratings or implicit feedback~\cite{nilashi2015multi,zhangCausalNeuralFuzzy2021}; and (iv) augment a recommender’s prediction with outputs from a fuzzy rule module~\cite{zhangCausalNeuralFuzzy2021,dengNovelFuzzyNeural2024}. These hybrids still rely on heuristic rule mining (e.g., Wang--Mendel), fixed rule sets, or mappings from data to predefined fuzzy atoms. In contrast, our model learns the rule structure and parameters end-to-end and then extracts human-readable rules at configurable thresholds.

\noindent\textbf{Our approach.} In this paper, we introduce transparent news recommender systems based on a Fuzzy Neural Network (FNN) architecture (similar to \cite{PEDRYCZ19931}) that learns human-readable rules from behavioral data to predict article clicks. As illustrated in Figure~\ref{fig:architecture}, the model composes fuzzy \texttt{AND}/\texttt{OR}/\texttt{NOT} operations in a layered architecture, enabling rich, expressive rules. A simple thresholding procedure extracts interpretable rules whose complexity is \emph{configurable}: by tuning the threshold, practitioners can trade off rule length (and thus interpretability) against predictive accuracy. Beyond explanations, these rules reveal consumption patterns that can be aligned with editorial goals, providing a practical handle for human oversight.
 
\noindent We conduct offline experiments on two publicly available news datasets, i.e., MIND and EB-NeRD. The MIND~\cite{DBLP:conf/acl/WuQCWQLLXGWZ20} dataset contains the user click logs and article features of Microsoft News, a news aggregator in English. 
EB-NeRD~\cite{kruse2024eb} contains clicks of the Danish newspaper \emph{Ekstra Bladet}\footnote{\url{https://ekstrabladet.dk/}}, as well as additional user and article features. %
We find that our model accurately predicts click behavior compared to several established baseline methods while producing interpretable, human-readable rules. We further show how the extracted rules capture news consumption dynamics (e.g., recency effects), illustrating how our approach can support editors in curating news content. \\

\noindent Summing up, we make the following contributions\footnote{The code is available at \url{https://anonymous.4open.science/r/FNN4NewsRecommendation/}}:
\begin{itemize}[topsep=0pt]
    \item We propose an architecture for a Fuzzy Neural Network (FNN) with multiple layers, negation, and regularization. %
    \item We demonstrate the efficacy of our approach on two news datasets, i.e., MIND and EB-NeRD, in relation to several baselines. %
    \item We extract rules from our model's weights and show the alignment of the extracted rules with the model's predictions. %
    \item We analyze news reader behavior, based on the extracted rules of our FNN, and discuss the application of our model as a support tool for news editors.
\end{itemize}

\section{Interpretable Rule Learning with Fuzzy NNs}
\label{sec:methodology}

We propose a Fuzzy Neural Network (FNN) that composes fuzzy logic operators in a layered architecture. Compared to prior work~\cite{bartl2025difffuz}, our approach supports multiple layers, negation operations, and includes orthogonality regularization~\cite{2018arXiv181009102B} to improve rule interpretability.

\para{Fuzzy Neural Network}
\label{ssec:fnn}
Let $L$ be the number of layers and $\mathbf{o}^{(0)}\in[0,1]^{n_0}$ the input atoms. Each layer consists of nodes performing fuzzy logic operations based on the product T-norm~\cite{klement2013triangular}, defined
for two fuzzy variables $a,b \in [0,1]$ as:
\begin{align*}
\texttt{AND}(a, b) \coloneq a \cdot b,\quad
\texttt{OR}(a, b) \coloneq a + b - a \cdot b,\quad
\texttt{NOT}(a) &\coloneq 1 - a
\end{align*}
All nodes in a layer compute the same operation (\texttt{AND} or \texttt{OR}).
Similar to a multi-layer perceptron, the output of each layer $\mathbf{o}^{(l)}\in[0,1]^{n_{l}}$ (for $1\leq l\leq L$) is computed from the output of the previous layer $\mathbf{o}^{(l-1)}$ and a fuzzy weight matrix $\mathbf{W}^{(l)}\in[-1,1]^{n_{l}\times n_{l-1}}$. The weighted inputs $\mathbf{Z}^{(l)}\in[0,1]^{n_{l}\times n_{l-1}}$ are defined as $\mathbf{Z}^{(l)}_{ji}=\texttt{OR}(\mathbf{o}^{(l-1)}_i, 1-|\mathbf{W}^{(l)}_{ji}|)$ for \texttt{AND} layers and as $\mathbf{Z}^{(l)}_{ji}=\texttt{AND}(\mathbf{o}^{(l-1)}_i, |\mathbf{W}^{(l)}_{ji}|)$ for \texttt{OR} layers~\cite{PEDRYCZ19931}. When the weight is negative $\mathbf{W}^{(l)}_{ji}<0$ we negate the input before weighting, i.e., $\texttt{NOT}(\mathbf{o}^{(l-1)})$.
The layer output is then defined as $\mathbf{o}^{(l)}_j = \prod_{i=1}^{n_{l-1}} \mathbf{Z}^{(l)}_{ji}$ for an \texttt{AND} layer and as $\mathbf{o}^{(l)}_j = \texttt{NOT}\big(\prod_{i=1}^{n_{l-1}}\texttt{NOT}(\mathbf{Z}^{(l)}_{ji})\big)$ for an \texttt{OR} layer.
The weights define how strongly a fuzzy input contributes to the logical conjunction (or disjunction); a weight of $0$ means no contribution at all, a weight of $1$ means full contribution (as in crisp logic).

\noindent The fuzzy weights $\mathbf{W}^{(l)}$ are derived from the model parameters $\mathbf{P}^{(l)}\in\mathbb{R}^{n_l}$ to enable unconstrained gradient descent. Unlike prior work~\cite{bartl2025difffuz}, we apply a $\tanh$ function to $\mathbf{P}^{(l)}$, constraining the weights to the interval $[-1,1]$. 
This modification allows the model to learn negations---as described above---, hence enabling the representation of any logical formula. %

\para{Discretization of features}
\label{ssec:discretization}
As fuzzy logic requires values in $[0,1]$, we apply several preprocessing steps.
\emph{Categorical features} with $n$ categories are transformed to crisp sets $v=\{0,1\}^{21}$ using one-hot encoding covering the 20 most frequent categories plus a category \texttt{infrequent}.
\emph{Multi-class features} are converted to crisp sets; multiple entries can be $1$ and all categories are kept.
\emph{Numerical features} are discretized into three quantile-based bins: \texttt{low}, \texttt{mid} and \texttt{high}. %
\emph{Missing values} are replaced by the mean of the corresponding fuzzy set.

\para{Rule Extraction}
\label{ssec:rule_extraction}
\RestyleAlgo{boxruled}
\LinesNumbered
\begin{wrapfigure}[21]{R}{0.67\textwidth} %
\vspace{-1.0cm}
\begin{minipage}{.67\textwidth}
\begin{algorithm}[H]
\caption{A rule extraction algorithm. Removes all connections with a weight below the threshold $t$, thus mapping the fuzzy logical operations to crisp logical operators.}\label{alg:rule_extraction}
\scriptsize
\KwIn{$W\ldots$Trained FNN, $a_i\ldots$Symbol for the $i$-th fuzzy input atom, $t\ldots$Extraction threshold}
$r \gets [a_i | i\in[1,2, \ldots]]$ \Comment*[r]{Init clauses with input atoms}
\For{$l\in [1, 2, \ldots, L]$}
{
    $\mathbf{W}^{(l)}\in[0,1]^{\mathrm{out}\times \mathrm{in}}$ \Comment*[r]{Weights of current layer}
    $r'\gets[~]$ \Comment*[r]{The new clauses}
    \For{$o\in [1, 2, \ldots, \mathrm{out}]$}
    {
        $c\gets[~]$ \Comment*[r]{A new clause}
        \For{$i\in [1, 2, \ldots, \mathrm{in}]$}
        {
            \Comment{Store only clauses with high weights}
            \If{$|\mathbf{W}^{(l)}_{o,i}| > t$}
            {   
                $c\gets c + [r_i]\text{ if }\mathbf{W}^{(l)}_{o,i} > 0\text{ else }c + [\texttt{NOT}(r_i)]$ 
            }
        }
        $\bigoplus\in\{\bigwedge, \bigvee\}$ \Comment*[r]{Logic operator of layer $l$}
        $c\gets \bigoplus_{k=1}^{\mathrm{length}(c)} c_k$ \Comment*[r]{ignore $c_k=[~]$}
                $r'\gets r' + [c]$ \Comment*[r]{Append the new clause}
    }
    $r\gets r'$ \Comment*[r]{Store new clauses for next iteration}
}
\Return{$r_0$} \Comment*[r]{The final clause is the extracted rule 
}
\end{algorithm}
\end{minipage}
\end{wrapfigure}
To extract rules for analysis and interpretation, we use thresholding on the weights. Connections with $|\mathbf{W}^{(l)}_{ji}|<t$ are pruned, while the remaining connections are mapped to a crisp logical formula. Starting at the input atoms, we construct OR/AND clauses based on the previous layer’s output to represent the internal logic of our FNN (as outlined in Algorithm~\ref{alg:rule_extraction}). %

\section{Learning News Behavior for Editorial Oversight}
\label{sec:news_domain}

We evaluate our approach on the small versions of the EB-NeRD~\cite{kruse2024eb} and MIND~\cite{DBLP:conf/acl/WuQCWQLLXGWZ20} datasets, using the provided temporal train-validation splits.
Our model is trained to classify whether an item is relevant to a user, given fuzzy features about both the user and the article. %
While this task is not directly aligned with the goal of ranking news articles, we can use the fuzzy model predictions to rank articles.

\subsubsection{Article Age Atoms}
The relevance of news articles is inherently time-sensitive, as user engagement declines over time~\cite{DBLP:conf/acl/WuQCWQLLXGWZ20}. 
To capture temporal dynamics, we encode \emph{article-age} %
with the following time bins:
\texttt{$30$min}, \texttt{$1$h}, \texttt{$2$h}, \texttt{$4$h}, \texttt{$8$h}, \texttt{$12$h}, \texttt{$1$d}, \texttt{$2$d}, \texttt{$3$d}, and \texttt{$>3$d}.
We choose to put more focus on the recent timescale, and stop binning if more than $3$ days passed.
Each article is only assigned to the bin with the smallest article age, so a $3$ hour old article belongs \emph{exclusively} to the \texttt{$4$h} bin.

\subsubsection{EB-NeRD Atoms}

For our experiments, we leverage user and article information given in the EB-NeRD dataset. %
Specifically, we use \emph{device type}, users' \emph{gender}, their \emph{age} (in 10-year bins), their district category based on its \emph{postcode}, and whether they are a \emph{subscriber} for premium content ($3.35$\% of users).
Data on \emph{gender}, \emph{age}, and \emph{postcode} is sparse, with over 95\% of entries ``unknown''.
For articles, we encode the \emph{article-type}, \emph{category}, \emph{sub-category}, \emph{topics}, \emph{sentiment-label}, and \emph{premium} subscription.
Additionally, we model behavioral data from the first seven days of publication concerning the total number of times it was shown  (i.e., \emph{in-views}), clicked (i.e., \emph{page-views}), and the total time read (i.e., \emph{read-time}).
The \emph{article-type} and \emph{category} both show skewed distributions: 4 of the 12 \emph{article-types} and 9 of the 25 \emph{categories} each account for 99\% of the articles.

\subsubsection{MIND Atoms}

Unlike the EB-NeRD dataset, the MIND dataset offers only 
limited features for both user and articles. %
For users, only the ID and the history is available, which restricts our ability to model user attributes. 
For articles, we encode the \emph{category}, the \emph{sub-category}, detected \emph{entities} in the article title and abstract (using named entity recognition by the dataset creators). Common entities are ``Donald Trump'', ``United States'' and ``National Football League'',
showing a strong focus on US news in the  dataset.

\para{Hyperparameters}
\label{ssec:training}
We use \textbf{AdamW}~\cite{DBLP:conf/iclr/LoshchilovH19} ($lr=0.01$, $\mathrm{batchsize}=8196$, $\mathrm{epochs}=20$, weight decay $=10^{-5}$) with binary cross-entropy loss (BCE).
To direct the model towards simpler rules, we use L1 regularization ($\lambda_{L1}=\frac{0.1}{\#\mathrm{parameters}}$). Moreover, we apply double soft orthogonal regularization~\cite{2018arXiv181009102B} ($\lambda_{orth}=\frac{0.1}{\#\mathrm{parameters}}$) to encourage different clauses within a layer.
The architecture includes three hidden layers ($48, 32, 16$ neurons) and a single output layer with alternating \texttt{AND}/\texttt{OR} operators between the layers.
We evaluate two settings: \textbf{\mbox{FNN-and}} (output node is \texttt{AND}) and \textbf{\mbox{FNN-or}} (output node is \texttt{OR}).
The model's weights are initialized with Kaiming uniform initialization~\cite{DBLP:conf/iccv/HeZRS15} and a gain of $\frac{5}{3}$.
As users typically click on fewer articles than shown, we randomly sample $K=4$ non-clicked articles per impression~\cite{DBLP:conf/kdd/WuWAHHX19}.

\para{Evaluation}
\label{ssec:evaluation}
Following prior work on the datasets, we compute AUC, MRR, nDCG@5, and nDCG@10~\cite{DBLP:conf/acl/WuQCWQLLXGWZ20,kruse2024eb} on the item rankings, grouped by impression.
Moreover, we introduce a custom metric, ``Rule Complexity'' $\mathrm{RC}@t$, defined as the number of logical operators, i.e., $\land$, $\lor$, and $\lnot$, in the crisp rule, extracted at threshold $t$ (see Subsection~\ref{ssec:rule_extraction}). 

\noindent We compare our approach against the following baselines:
\emph{Random}, which predicts random scores for user-item pairs with a truncated ($[0,1]$) normal distribution (estimated from the train set);
\emph{User-KNN} and \emph{Item-KNN}, both collaborative k-nearest-neighbor models based on user-user and item-item similarities;
\emph{SVD++}, which performs singular vector decomposition of the user-item matrix, with a modification for implicit feedback~\cite{DBLP:conf/kdd/Koren08};
\emph{Decision Tree}, trained on the crisp atoms, which learns purely crisp rules in an interpretable tree structure.
\emph{NRMS}, an established two-tower neural model for news recommendations~\cite{wu2019neural}.

\section{Results}
\label{sec:results}
We report the performance results on both EB-NeRD and MIND, while subsequent analyses focus on EB-NeRD, since it provides more features. 
For each experiment, we report the mean result across 5 runs.

\subsection{Quantitative Results}
\label{ssec:quantresults}

We conducted quantitative experiments on the two datasets averaged over five runs (see Table~\ref{tab:quant_results}). We evaluate the predictive power by comparing against six baselines (including one deep-learning black-box model, i.e., NRMS, without interpretability features). %
KNN and SVD achieve low performance in terms of all metrics, except AUC, due to the cold-start problem. This is due to the temporal split and articles losing relevance over time. Hence, many items were not included in the training set and KNN and SVD instead predict the global mean score. %

\begin{table}[htpb]
\caption{Quantitative results (vs. baselines). Best: \textbf{Bold}, 2\textsc{nd} Best: \underline{Underlined}.}
\label{tab:quant_results}
\centering
\begin{tabular}{l|cccc|cccc}
\toprule
\multirow{2}{*}{Method} & \multicolumn{4}{c}{\textbf{EB-NeRD Small}} & \multicolumn{4}{|c}{\textbf{MIND Small}} \\
 & AUC & MRR & nDCG@5 & nDCG@10 & AUC & MRR & nDCG@5 & nDCG@10 \\
\midrule
Random & .5002 & .3015 & .3211 & .4119 & .4996 & .2184 & .2228 & .2858 \\
User-KNN & .5005 & .1681 & .1385 & .2716 & .5004 & .2192 & .2242 & .2864 \\
Item-KNN & .5006 & .1669 & .1381 & .2712 & .4998 & .2183 & .2235 & .2858 \\
SVD++ & .5010 & .3059 & .3382 & .4244 & .5003 & .2197 & .2243 & .2871 \\
Decision Tree & .6039 & .3784 & .4164 & .4840 & \underline{.5812} & \underline{.2717} & \underline{.2963} & \underline{.3532} \\
\midrule
NRMS & .5548 & .3488 & .3878 & .4648 & \textbf{.6548} & \textbf{.3094} & \textbf{.3402} & \textbf{.4043} \\
\midrule
FNN-or (ours) & \underline{.6915} & \textbf{.4582} & \underline{.5175} & \underline{.5662} & .5766 & .2450 & .2716 & .3339 \\
FNN-and (ours) & \textbf{.6925} & \underline{.4580} & \textbf{.5176} & \textbf{.5664} & .5806 & .2485 & .2752 & .3355 \\
\bottomrule
\end{tabular}
\end{table}

\noindent On EB-NeRD, our method outperforms all baselines in terms of all metrics. 
Still, better performing models have been reported previously~\cite{DBLP:conf/recsys/KruseLKPP0UAF24a} (up to 79.10 for nDCG@10 on testset). Nonetheless, we find that our model provides accurate predictions, while being small with $2064$ ($48\times32+32\times16+16\times1$) parameters.

\noindent On the MIND dataset, we observe lower performance. One reason is the lack of usable features for our model input (see Section~\ref{sec:news_domain}). This reflects a limitation of our approach: performance depends on the richness of provided features. Although we can derive additional signals from articles, e.g., the article age (see Section~\ref{ssec:fnn}), this still depends on the dataset. 

\noindent While we outperform the decision tree, which uses the identical input as our model, i.e., the discretized features, on EB-NeRD, the performance on the MIND dataset is lower. This suggests that while our model's performance scales well with more features, when the model complexity stays the same, a decision tree can be competitive (or better) when the available input features are scarce.

\subsection{Ablation Study}
\label{ssec:ablat}

\begin{table*}[t]
\caption{Ablation study results showing the mean metrics when using different components for our model.
The star (*) marks significant ($\alpha=0.05$) differences in a paired Dunn's test with Bonferroni correction with the FNN-and config.}
\label{tab:ablation-study}
\begin{tabular}{l|cccc|rrr}
\toprule
Experiment & \textbf{AUC} & \textbf{MRR} & \textbf{nDCG@5} & \textbf{nDCG@10} & \textbf{RC@0.4} & \textbf{RC@0.5} & \textbf{RC@0.6} \\
\midrule
FNN-and (ours) & \textbf{.692}\phantom{*} & \underline{.457}\phantom{*} & \textbf{.517}\phantom{*} & \underline{.565}\phantom{*} & 175.2\phantom{*} & 91.2\phantom{*} & 27.4\phantom{*} \\
FNN-or (ours) & .689\phantom{*} & .456\phantom{*} & .514\phantom{*} & .564\phantom{*} & 48.4\phantom{*} & 13.2\phantom{*} & 3.4\phantom{*} \\
\midrule
BCE $\rightarrow$ MSE Loss & \underline{.691}\phantom{*} & \textbf{.458}\phantom{*} & \textbf{.517}\phantom{*} & \textbf{.566}\phantom{*} & 70.8\phantom{*} & 33.8\phantom{*} & 13.4\phantom{*} \\
w/o L1 Reg. & \underline{.691}\phantom{*} & .456\phantom{*} & \underline{.516}\phantom{*} & .565\phantom{*} & 67.8\phantom{*} & 40.8\phantom{*} & 16.6\phantom{*} \\
w/o Orthog. Reg. & .687\phantom{*} & .453\phantom{*} & .512\phantom{*} & .562\phantom{*} & 5160.4\phantom{*} & 2475.0\phantom{*} & 627.0\phantom{*} \\
w/o Neg. Sampling & .680\phantom{*} & .448\phantom{*} & .504\phantom{*} & .555\phantom{*} & 77.0\phantom{*} & 41.0\phantom{*} & 26.4\phantom{*} \\
w/o Article Age & .600* & .374* & .418* & .491* & 184.8\phantom{*} & 80.0\phantom{*} & 28.0\phantom{*} \\
Tanh $\rightarrow$ Sigmoid & .523* & .310* & .343* & .429* & 1.0* & 1.0* & 1.0\phantom{*} \\
\bottomrule
\end{tabular}
\end{table*}

To validate our components, we perform an ablation study (Table~\ref{tab:ablation-study}) altering the model configuration: (i) ``BCE $\rightarrow$ MSE Loss'', uses mean squared error (MSE) instead of BCE, (ii) ``w/o L1 Reg.'' and ``w/o Orthog. Reg.'', remove the L1 loss and orthogonality loss terms, (iii) ``w/o Neg. Sampling'' use all available interactions instead of negative sampling, while ``w/o Article Age'' removes the article age input, and (iv) ``Tanh $\rightarrow$ Sigmoid'' uses sigmoid for the weight computation, which removes the model's capabilities to model negations. We observe significant differences between runs for all metrics using Kruskal-Wallis H-test, %
and perform a pairwise Dunn's test with Bonferroni correction per metric.

Switching to MSE does not decrease performance, and slightly improves two metrics, but consistently yields lower weights. 
For interpretability, this matters: rule extraction thresholds a fuzzy model into crisp rules, and the usefulness of this improves when the model's weights are either $1$ or $0$, meaning the extracted crisp rule---for any threshold, except $t=0$ and $t=1$---would perfectly describe the model's logic. Because BCE better reflects this than MSE in our setting, we keep BCE as the default. Finally, using a sigmoid for the weight computation removes the ability to represent negations, reducing the model's expressiveness.

\subsection{Sensitivity Analysis}
\label{ssec:sensitivity}

\begin{figure*}[t!]
\centering
\begin{subfigure}[t]{0.5\textwidth}
    \centering
    \includegraphics[width=\linewidth]{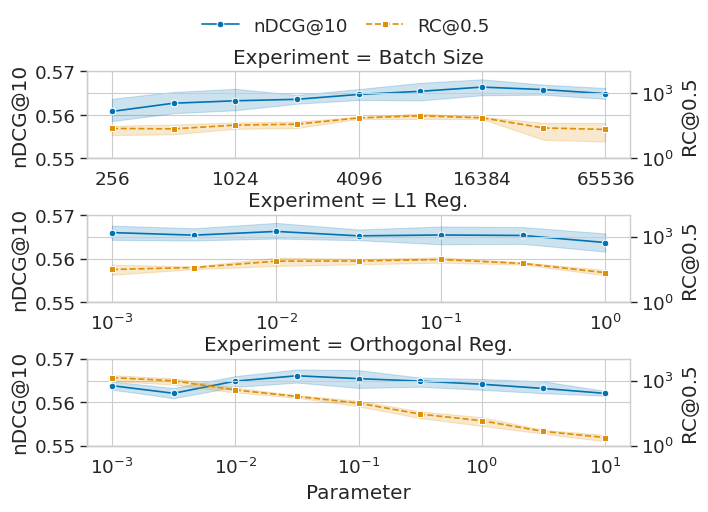}
    \caption{The sensitivity analysis of the batch size, the L1 regularization and the orthogonality regularization. We show both the prediction performance, with nDCG@10, and the rule complexity with RC@0.5. The parameters (x-axis) and RC@0.5 (right y-axis) are on logarithmic scale. 
    }
    \label{fig:sensitivity}
\end{subfigure}%
~ 
\begin{subfigure}[t]{0.5\textwidth}
    \centering
    \includegraphics[width=\linewidth]{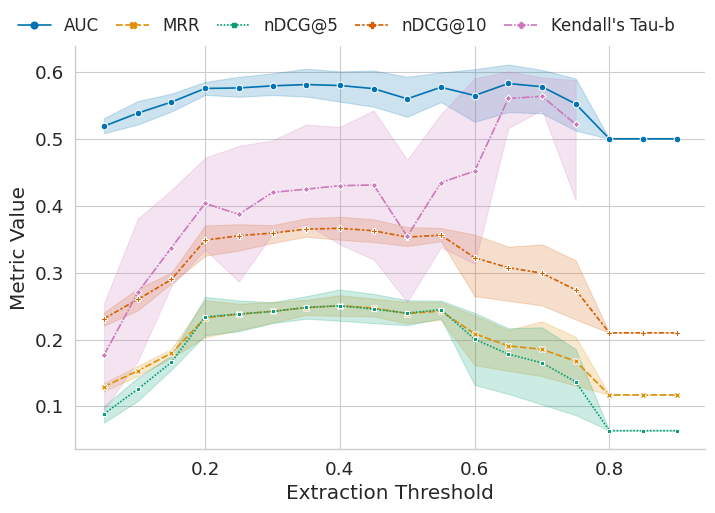}
    \caption{Performance of extracted crisp rule on the EB-NeRD-small validation split, visualized for multiple threshold values.
    Additionally, the correlation of the predicted rankings with the model's predicted rankings is measured using Kendall's Tau-b rank-based correlation metric.}
    \label{fig:rule_extraction}
\end{subfigure}
\caption{Analysis of parameter sensitivity (training) and threshold (prediction).}
\end{figure*}

We analyze the sensitivity with %
a parameter sweep over the batch size, the L1 regularization, and the orthogonal regularization, and visualize prediction performance and rule complexity in Figure~\ref{fig:sensitivity}. %

For the batch size, we want to highlight the region between 4096 and 16384. Here, the rule complexity varies the least between runs, while the accuracy stays high. The default batch size of 8096 falls right in the middle of this range, and although the prediction performance is slightly worse than for 16384, the error bars across runs (95\% confidence interval) largely overlap.

The L1 regularization does not have a strong impact on model performance, but we can see that we get the most expressive rules at $10^{-1}=0.1$. We can also see a trend towards less expressive rules, at a threshold of 0.5, for both high and low L1 regularization values. In the ablation study (Section~\ref{ssec:ablat}), we observed that when turning off L1 regularization completely, rule complexity decreases to about half. This might seem counterintuitive, since L1 loss encourages feature selection. The feature selection pushes most of the weights close to 0, instead focusing on a few inputs, for which the weights will be pushed towards 1. This, therefore, leads to more rules being extracted, because important features have higher weights, specifically, weights higher than the threshold of 0.5. The chosen default L1 regularization parameters is therefore $0.1$.

The orthogonal regularization has a large impact on the rule complexity. We can observe a clear log-log relationship between the parameter and the rule complexity at threshold 0.5. Since we want good prediction performance, but also want rules which are expressive, while still being interpretable, we choose the default parameter as $0.1$. Note that this parameter is useful to control the complexity extracted rules, though with a small impact on prediction performance.

\subsection{Interpretation of Rules}
\label{ssec:interpretrules}

Figure~\ref{fig:rule_extraction} shows the performance of the extracted rules at various thresholds, keeping the model's weights frozen.
Since our extracted rules are crisp, the only possible outputs for an impression are \textbf{true} and \textbf{false}, which in a ranking task produces ties for all \textbf{true} impressions and all \textbf{false} impressions.

To validate the alignment of the extracted rules with the model's actual predictions, we compute the correlation of the global ranking of the model and the extracted rules. For the correlation statistic, we use Kendall's Tau-B, which adjusts for ties. We observe a positive trend for the correlation with increasing threshold levels, while accuracy saturates between $0.2$ and $0.55$. %

For thresholds $>0.8$, the extracted rule predicts a constant output, \textbf{true} or \textbf{false}, either because the rule is a contradiction, e.g., $a\land\neg{a}$, or a tautology, e.g., $a\lor\neg{a}$. Since the predictions are all identical, Kendall's Tau-B is not computable.

\subsection{Qualitative Analysis of the Extracted Rules}
\label{ssec:qualianalysis}

We extract crisp rules using a threshold of $0.55$, which gives us rules with a low rule complexity and a high nNDG. To further improve rule readability, we group article age time-spans, if they are consecutive, e.g., 2h-4h and 4h-8h as 2h-8h.

We treat the input of the last layer as distinct rules and keep their weights; this allows for increased interpretability. Since we investigate the FNN-and configuration here, where the last node is an \texttt{AND}, the rules are \emph{requirements} for relevance. In case the final node is an \texttt{OR}, the rules are \emph{indicators} of relevance.

The extracted rules in Table~\ref{tab:qualitative_analysis} show that article relevance depends primarily on \emph{article age} or the ``page9'' category, represented by $C_\texttt{side9}$.
The boolean symbol $C_\texttt{nationen}$ corresponds to the ``nations'' category, while
$SC_\texttt{265}$ depicts an undisclosed subcategory with ID 265. $AA_\text{interval}$ indicates that the article's age lies in the ``interval''. %
$IV_\texttt{low}$ states that the in-view count belongs to the low quantile. 

The highest-weighted rule (R1) requires articles younger than one day, except for those in the ``page9'' category. Rule R2---and the identical rule R3---is always \textbf{true}, because the article age disjunction term is always \textbf{true}, i.e., $\neg{}AA_{30m-1h} \lor{} \neg{}AA_{1d<}$. Rule R4---identical to rule R5---needs article published less than 30 minutes or more than 1 hour ago ($AA_{2h-1d}$ lies within this range, and can be ignored for the crisp rules).
Finally, R6 is technically always \textbf{true} ($\neg{}AA_{30m-1h} \lor{} \neg{}AA_{1d<}$), but also shows that our model uses the ``nations'' category, subcategory 265 and low in-view counts to predict user clicks.

\begin{table}[t]
\caption{Extracted rules from the FNN-and output node.}
\label{tab:qualitative_analysis}
\centering
\begin{tabular}{l|r|l}
\toprule
\textbf{Rule} & \textbf{Weight} & \textbf{Extracted Crisp Rule} \\
R1 & 0.95 & $C_\text{side9} \lor{} \neg{}AA_{1d<}$ \\
R2 & 0.94 & $C_\text{side9} \lor{} \neg{}AA_{30m-1h} \lor{} \neg{}AA_{1d<}$ \\
R3 & 0.93 & R2 \\
R4 & 0.83 & $\neg{}AA_{30m-1h} \lor{} AA_{2h-1d}$ \\
R5 & 0.66 & R4 \\
R6 & 0.57 & $C_\text{nationen} \lor{} SC_\text{265} \lor{} IV_\text{low} \lor{} \neg{}AA_{30m-1h} \lor{} \neg{}AA_{1d<}$ \\
\bottomrule
\end{tabular}
\end{table}

\section{Theory of Change: A transparent, fuzzy neural news recommender for editorial oversight}
We address the need for transparency by design in news recommendation, where black-box news recommender systems limit editorial oversight and conflict with rising algorithmic transparency mandates as posed by, for example, the EU AI Act~\cite{schedl2025technical}. Our approach is intended not as a replacement for editors, but as a decision-support tool (rather than for news delivery automation). Our news recommender is trained on impression logs as well as user and article metadata, and outputs crisp, human-readable rules reflecting patterns observed in reader behavior. These rules, quantified by Rule Complexity (RC@t) and Kendall’s $\tau_b$, correspond to the outputs. In this way, we enable editors to diagnose underlying reasons for recommendations and govern recommendations by editing rules or adjusting thresholds, thereby adapting algorithmic behavior to journalistic values and editorial policies. The intended impact is transparency and the support of editorial responsibility in news curation. This theory of change assumes access to log data, editorial engagement and processes that allow for diagnostics. Risks include oversimplification, technology avoidance in the editorial room~\cite{moller2022recommended}, and data sparsity; we mitigate them by exposing the threshold $t$, reporting Rule Complexity (RC@t) and Kendall’s $\tau_b$, and by monitoring ranking performance. %

\section{Conclusion and Future Work}
\label{sec:conclusion}
Our FNN model accurately predicts click behavior, while exposing news consumption patterns relevant for editorial oversight.
We validate our approach through an ablation study and a sensitivity analysis. Furthermore, we confirm that the extracted rules align with the model's predictive logic, and provide a qualitative analysis of these extracted rules in the context of news consumption.
Such information can support editors in their content curation goals, such as balancing recency in news coverage. 
For future work, we plan to investigate how these learned rules could be integrated with manually defined rules, such as editorial guidelines. Besides, we plan to make our model decomposable, smooth, and deterministic (as defined in~\cite{DBLP:journals/jair/DarwicheM02}) to allow for tractable queries (e.g., which topics are more relevant for a certain demographic), thus enabling auditing capabilities to adhere to regulations.

\begin{credits}
\subsubsection{\ackname} This research was funded in whole or in part by the Austrian Science Fund (FWF) 10.55776/COE12.

\end{credits}

\appendix

\bibliographystyle{splncs04}
\bibliography{main}

\end{document}